\pdfoutput=1

\documentclass[11pt]{article}

\usepackage{ACL2023}

\usepackage{times}
\usepackage{latexsym}
\usepackage[T1]{fontenc}

\usepackage[utf8]{inputenc}

\usepackage{microtype}

\usepackage{inconsolata}

\usepackage{svg}
\graphicspath{ {./images/} }
\usepackage[title]{appendix}
\usepackage{xurl}

\title{Linguistic ambiguity analysis in ChatGPT}

\author{
Miguel Ortega-Martín\\
  dezzai, UCM \\
  \small\texttt{m.ortega@dezzai.com} \\
  \small\texttt{m.ortega@ucm.es} \\
  \And
  Óscar García-Sierra \\
  dezzai, UCM \\
  \small\texttt{oscar.garcia@dezzai.com} \\
  \small\texttt{oscarg02@ucm.es} \\
  \And
  Alfonso Ardoiz \\
  dezzai \\
  \small\texttt{alfonso.ardoiz@dezzai.com} \\
    \AND
   \textbf{Jorge Álvarez} \\
  dezzai \\
  \small\texttt{jorge.alvarez@dezzai.com} \\
  \small\texttt{} \\
  \And
  \textbf{Juan Carlos Armenteros} \\
  dezzai \\
  \small\texttt{juancarlos.armenteros@dezzai.com} \\
  \And 
  \textbf{Adrián Alonso} \\
  dezzai, URJC, DSL URJC\\
  \small\texttt{a.alonso@dezzai.com} \\
  \small\texttt{adrian.barriuso@urjc.es} \\
  }

\begin{document}

\maketitle

\begin{abstract}

Linguistic ambiguity is and has always been one of the main challenges in Natural Language Processing (NLP) systems. Modern Transformer architectures like BERT, T5 or more recently InstructGPT have achieved some impressive improvements in many NLP fields, but there is still plenty of work to do. Motivated by the uproar caused by ChatGPT, in this paper we provide an introduction to linguistic ambiguity, its varieties and their relevance in modern NLP, and perform an extensive empiric analysis. ChatGPT strengths and weaknesses are revealed, as well as strategies to get the most of this model.

\end{abstract}

\section{Introduction}

Sometimes on a day-to-day basis humans struggle to solve linguistic ambiguity, so we must accept that maybe machines may also face difficulties alike. As \citet{jurafsky20} state at the start of one the most important Natural Language Processing books, most of NLP tasks could be seen as a disambiguation task in any of the six levels in which linguistic is built: phonetics, morphology, syntax, semantics, pragmatics and discourse. In the example sentence given by \citet{jurafsky20}, "I made her duck", we can encounter as much as five different senses, which would require disambiguating in most of the six previous levels. In this work we will focus on three of them, how they are dealt with in modern NLP systems and what room is still there for improvement. 

This work contains a general review of ambiguity in linguistics (section 2); a review  of how modern NLP systems handle ambiguity problems (section 3); an introduction to ChatGPT (section 4); an empirical analysis of different types of linguistic ambiguity with ChatGPT (section 5); the conclusions (section 6); the limitations of our work (section 7); and the Ethics Statements (section 8). Our contributions in this paper are the following:

\begin{itemize}
    \item review of current challenges of linguistic ambiguity in NLP.
    \item analysis of the linguistic weaknesses and strengths of the ChatGPT model related to linguistic ambiguity.
\end{itemize}

\section{Linguistic Ambiguity}

Typically, three types of linguistic ambiguity are reviewed when analysing modern NLP systems: lexical, syntactic and semantic. More complex levels, like pragmatics and discourse, also present ambiguities, but those are nowadays still way to far from being solved. 

We will focus on these three basic levels as they gather the main challenges faced by modern NLP models. However, sometimes the borders among levels are diffuse, and more than one kind of ambiguity could be applied to a given sentence. For instance, the lexical level also include ambiguities that happen in the phonetic level, and lexical and syntactic ambiguities mean a change of meaning and therefore can also be seen as semantic ambiguities. 

\subsection{Lexical Ambiguity}

Homonymy and polysemy are the two main representations of lexical ambiguity. Homonymy refers to those cases in which two words with different etymology and meaning end up having similar spellings or pronunciations. In the typical sentence "I went to the bank" the word "bank" can be a financial business or the area next to a river. Without further context, if we ask anyone "What bank refers to", its answer should be that the sentence is ambiguous.

As an additional note, we can talk about homophones (two words are read the same, but could be written differently) and homographs (two words are written and read the same).

On the other hand, polysemy occurs in those situations in which the same word can have several meanings. For example, the various meanings that the verb "to serve" has in the dictionary: "serve time in prison", "serve a tea", etc.

In conclusion, in homonymy, there are two words that have come to the same form. In polysemy, there is one word that has reached multiple meanings.  

\subsection{Syntactic Ambiguity}

This ambiguity does not come from a word, but from groups of words and their relationships. Also known as amphibology, syntactic ambiguity is due to an ambiguous sentence structure, in which a chunk or group of words in the sentence generates more than one interpretation of the whole sentence. A typical case of syntactic ambiguity is a type of structure like "Mary saw John with a telescope", in which the prepositional phrase "with a telescope" could imply that Mary used a telescope to see John, or that John was carrying a telescope when Mary saw him. 

\subsection{Semantic Ambiguity}

In the semantic ambiguity cases, a lack of context leads to more than one interpretation of a sentence. In the sentence "My mother and my sister were sad after she shouted at her" without further context, we cannot disambiguate to whom the pronouns "she" and "her" refer to. In other cases the lack of context could be bigger. In the sentence "She loves me" we do not have any clue of who "she" could refer to. Irony, sarcasm or metaphors are other cases hard to infer, since humans need knowledge of the world and language itself to understand any hidden meaning.

\section{Ambiguity in NLP}

Ambiguity is a common problem that cannot be solved by the majority of Language Models (LMs) used in NLP. Most of the time the main problem is the lack of context. From Information Retrieval (IR) and Question Answering (QA) to Text Summarization or Grammatical Error Correction systems \citep{jusoh18}, training Word-sense disambiguation (WSD) systems and knowledge bases (KB) are some of the techniques traditionally used \cite{bano15}.

\subsection{Lexical Ambiguity in NLP}

Traditionally, without further Part-of-Speech tagging, it was not possible to disambiguate homonym words in NLP tasks. For example, in the sentence "How does a bear bear with the pain?", it would be hard to differentiate the first, a noun, from the second, a verb. Static embeddings \citep{mikolov13} suffered this same problem, those two "bear" words would only have one same embedding. Yet modern contextual embeddings \citep{liu20} used in Transformers models made a huge advance as the embedding of these same words with different use would effectively be different. Same happens with polysemy: given two meaning of a word like the verb "to serve", they will have different contexts and, therefore, different embeddings.

This can be seen in the Entity Linking\footnote{\url{https://paperswithcode.com/task/entity-linking}} (EL) task, which is a common task where we need to disambiguate among personal names. For instance, in the sentence "Paris Hilton visited Paris" both "Paris" need to be considered as different entities and matched to their corresponding id from a KB. 

\subsection{Syntactic Ambiguity in NLP}

Syntactic ambiguity plays also a key role in NLP. For instance, dependency tagging is a common preprocessing technique in which every token of an input is given a syntactic tag with its corresponding root. Given an ambiguous syntactic structure, inferring one or another interpretation could have enormous relevance in latter tasks. 

SpaCy\footnote{\url{https://spacy.io/}} is a state-of-the-art Python library for NLP tasks, and Dependency tagging is among its capabilities. For the previous syntactically ambiguous sentence, "Mary saw John with a telescope", an spaCy Language Model (LM) for English\footnote{en\_core\_web\_lg model} labels the "telescope" as a complement of the subject, "Mary". In this case, having one or another interpretation will highly rely on how those structures were labelled in the training set of the model. If we introduce more context in the sentence (for instance, "Mary saw John with a telescope in his hand.") the ambiguity could be solved by a human being. However, as a proof of how complex is ambiguity still nowadays for NLP systems, even the best spaCy LM for English labels "telescope" as a complement of "Mary" in this sentence in which it is clearly a complement of "John". As stated before, mistakes like this, that could be due to the quality of the training data, could have a huge impact in latter tasks. Models rely on the quality of the training data and may also lead to confirmation bias due statistical relevance.

\subsection{Semantic Ambiguity in NLP}

\begin{figure}[htbp]
  \centering
  \includegraphics[scale=0.45]{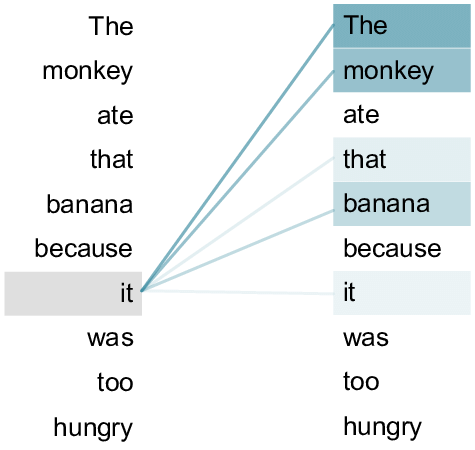}

  \caption{Attention mechanism. Source: \cite{xie21} }
\end{figure}

Semantic information and its disambiguation have also got better thanks to contextual embeddings and the attention mechanism. For instance, co-reference resolution has benefited from this advances. As seen in image 1, the pronoun "it" has a decent similarity with all the nouns in the sentence. However, attention is higher with "The" and "monkey", which allows us to disambiguate their co-reference.

\section{ChatGPT}

Large Language Models (LLMs) have taken over NLP. The current trend among them is to align their performance with the user intents. In 2022 OpenAI firstly released IntructGPT \citealp{ouyang22}, a fine-tuned model of GPT-3 \cite{brown20} with supervised learning and 1.3 billion parameter, in which human feedback plays a key role. At the end of 2022 OpenAI released a sibling model of IntructGPT, ChatGPT\footnote{\url{https://openai.com/blog/chatgpt/}}, another fine-tuned version of GPT-3, with 175 millions of parameters and trained to interact with users and receive feedback from them. 

In these models, prompting \citep{ouyang22} and human feedback play a huge role. A prompt is a piece of text used in the input that aims to reduce its context as it improves the quality of the output. 

Some work has quickly appeared\footnote{\url{https://medium.com/geekculture/i-spent-14-days-testing-chatgpt-here-are-3-ways-it-can-improve-your-everyday-life-30852a349ad1}} focusing on writing, automatizing, code generation, common knowledge question and reasoning. But, to the best of our knowledge, linguistic ambiguity has not been addressed yet.

Thanks to our experiments we have noted that by default it is preferable for these conversational models to give an answer, even if it is not of sufficient quality, rather than none at all. This can be applied to cases of poorly formulated or incomprehensible questions or to situations in which the model is not really capable of answering a well-formulated question.

\section{Experimental set-up and analysis}

To evaluate how ChatGPT deals with ambiguity, we have carried some experiments to measure its capabilities. At first, the only public dataset on ambiguity was considered: Sentence-Ambiguity on Big-BENCH \citep{srivastava_beyond_2022}. However, the type of ambiguity this test deals with, discourse level ambiguity, was not of interest to us. It was therefore necessary to create new tests to assess ambiguity at the lexical, syntactic and semantic levels. 

To do so, we input some ambiguous sentences via prompting and analyse ChatGPT outputs. In some occasions some queries required a change to another ChatGPT season, since we have realised that model's cache memory helps it to agree with the user in some situations without further reasoning. Appendixes contain all the sentences used in our empiric analysis. 

The main question for each sentence is if ChatGPT detects the presence or absence of ambiguity, but each kind of ambiguity will have its own particular target. In some cases, in addition to the initial ones, we have tried some other prompts that would allow us to go further in how ChatGPT works in general and to comprehend its relation with linguistic ambiguity in particular. For instance, in the case of homonymy it is relevant to ask if both words are the same, and in the case of polysemy, an special "definition classification" task is performed. 

\subsection{Lexical Ambiguity}

\subsubsection{Homonymy} 

Homonym research is focused on homographs (two words that are written and read the same) since, although homophones have a huge interest for Speech-to-Text tasks, they are not written the same, so it does not make sense to analyse them with a text-only system like ChatGPT. Our experiment is designed as the following:
\begin{itemize}
    \item We compile a list of sentences (Appendix A.1) and ask the model to label them as ambiguous or non-ambiguous with the following prompt: \textbf{Is the sentence "[sentence]" ambiguous?}
    \item Then we ask the model if the homonyms from a given sentence have the same meaning using the following prompt: \textbf{What does every occurrence of the word "[word]" mean?}
    \item Finally, in some cases we modified the original prompt to check any improvements in the outputs. 

\end{itemize}

Appendix A.2 contains all the inputs and outputs from this section. 

Before getting started with the experiment, human analysis is performed on the, in theory, non-ambiguous and previously mentioned sentence, "How does a bear bear with the pain?". This sentence contains words that need to be disambiguated in order to understand its meaning, but the full sentence does not seem to be ambiguous. The key point of it is that it contains two words "bear". The first one is a noun, and the second one is a verb. If we ask ChatGPT the question itself, as seen in example \ref{ex1.1} from Appendix A.2, the sentence is correctly understood, since it is, apparently, not ambiguous. 

Then, what happens if ChatGPT is explicitly asked if the previous sentence is ambiguous? As seen in query \ref{ex1.2}, this output contains some interesting points. In first place, ChatGPT does not find any ambiguity in the homonyms (both "bear"s), but does see polysemy in one of them, the verb "to bear", which makes the whole sentence ambiguous. This may be too excessive, since many people would not label this sentence as ambiguous just because of those two meanings of "to bear" (physically versus emotionally). This could be seen as one of those cases that some people has previously pointed, where ChatGPT have shown some overconfidence on its own answers. 

As we focus on homonymy, we also ask the model about the homonyms words themselves (query \ref{ex1.3} from Appendix A.2). With that simple query ChatGPT fails to detect both words. However, by just reformulating it, as in Query \ref{ex1.4}, the question gets solved, proving the relevance of correct prompting, and the model correctly detects homonymy. However, this case remarks the lack of consistency of ChatGPT, that in this case does not detect the previous polysemy of the verb "to bear".

The same results had been found on the remaining examples of homonymy (sentences 2-11 from Appendix A.1). As the previous one, these sentences are not intended to be ambiguous. For the list of ten sentences with a pair of homonyms in each one of them, just three out of the ten are labelled as non-ambiguous ("back", "can" y "band"). These false positive examples (queries \ref{ex1.5}-\ref{ex1.11} from Appendix A.2), which are relatively non-ambiguous for humans, allude to the polysemy of one of the homonyms, as shown in those outputs. 

In some occasions ChatGPT explanations look overconfident. Furthermore, in those explanations there seems to be more confusion when both words share Part-of-Speech tags ("bar", "right", "left"). In some other cases, confusion among nouns and verbs ("address", "rose", "book") makes us even question ChatGPT \textit{understanding} of syntax, since those words cannot be replaced by their homonym. 

Eventually, ChatGPT wrongly labels as ambiguous some sentences and confidently justifies a wrong explanation. Instead of a better ability detecting polysemy, ChatGPT muddles through as best it can -as humans grasp at straws when they struggle to find something that does not exist based keywords from the instructions. It is also worth noting that detecting more polysemous words than humans do is also problematic. If every word with more than on definition is labelled as polysemous, then half of the English dictionary would be polysemous. 

Results improve when we explicitly ask: "What does every occurrence of the word mean?". Outputs for sentences which are wrongly labelled as ambiguous due to polysemy according to the first question, amend and correctly detect words as homonyms according to this second question. However, there is one sentence ("bar") where both words are mistaken. 

Then we build three simple and clearly ambiguous sentences with some of the previous words (sentences 12-14 from Appendix A.1). We ask ChatGPT if these sentences are ambiguous, and the model labels just two of them as ambiguous, as seen in queries \ref{ex1.12}-\ref{ex1.14} from Appendix A.2. 

In conclusion, prompting is a key element. Doing the right question can achieve better results. Nonetheless, ChatGPT may show a wrong but convincing and elaborated output due to its overconfidence -as incorrect results for mathematical questions. Apparently ChatGPT tends to overdetect ambiguity, and its better at detecting positive ambiguous examples than negative ones. Also, ChatGPT brilliantly highlights the lack of context as the key element in a future disambiguation for some detected ambiguities. This last detail seems to indicate that the model has a default output assigned when it does not have sufficient information to generate a reliable response.

\subsubsection{Polysemy}

In this case, we use a sample of 10 lemmas (Appendix B). For each one of them:
\begin{itemize}
    \item We input a non-ambiguous sentence and tried to label it as ambiguous or non-ambiguous with the following prompt: \textbf{Is the sentence "[sentence]" ambiguous?}
    \item Then we design a meaning classification task, where we input two senses/definitions of the lemma and perform a classification task among those senses in the previous sentence. The prompt model is: \textbf{These are two definitions of "[lemma]": [DEF1, DEF2]. Which of the definitions appear in the sentence "[sentence]"?} 
    \item Also, for each sentence we ask ChatGPT to guess the meaning of the polysemous word, but without giving the definitions in the input, as in the second experiment. The prompt model is: \textbf{In the sentence "[sentence]" which is the definition of "[lemma]"?}
    \item In fourth place we create a collection of clearly ambiguous sentences using those lemmas, and try to label them as ambiguous or non-ambiguous with the following prompt: \textbf{Is the sentence "[sentence]" ambiguous?}
    \item  Finally, in the last experiment we try to confuse ChatGPT using erroneous pairs of lemma-definitions.

\end{itemize}

Appendix B.2 contains all the inputs and outputs from this section. 

In the first experiment we want to test whether ChatGPT recognises the concept of ambiguity, and whether it is able to detect it in simple sentences. The result is that ChatGPT correctly detects non-ambiguity in all of the ten sentences. Since a classification task must be done on them, all sentences are suppose not to be ambiguous, so ChatGPT does a great job by not detecting ambiguity. 

For the second experiment we design a meaning classification task, where for each of the ten lemmas we input two senses/definitions of each and one sentence to perform a classification task among those senses. Definitions and sentences come from Cambridge Dictionary\footnote{\url{https://dictionary.cambridge.org/}}. For instance, this is an example input for this second experiment: "These are two definitions of "to serve": 1. to provide food or drinks 2. to work for; to do your duty to. Which of the definitions appear in the sentence "She has served on the committee for the last 15 years"?"

For this second test, ChatGPT correctly classifies nine out of ten sentences. It only fails to classify the verb "to cheat", as seen in query \ref{ex2.1} from Appendix B.2. However, this can be solved with another better prompt in which the definitions are removed (query \ref{ex2.2} from Appendix B.2). In this case, less context is better. In order to go further with this idea, in the third experiment we check what happens if no definitions context is given. In the other nine sentences, ChatGPT easily gives a correct definition for all the nine polysemous words. 

As it was done in the case of homonymy, it's important to test clearly ambiguous examples. Sentences like "I saw the mouse on the table" and "I saw the mouse on the desktop" could have one predominant interpretation, but it is not either 100\% clear for humans. However, ChatGPT does not see any ambiguity, and clearly associates the first one with the animal and the second one with the electronic device. Given how easily the model previously detected polysemy, it is shocking that it does not detect it here. 

Same happens in the sentence "I spent the morning in the bank", where ChatGPT clearly identifies the first definition. In our opinion, there is not enough context and they should be flagged for ambiguity.

Following the idea of trying to improve prompting, if we ask ChatGPT if those two sentences with the term "mouse" are ambiguous, it labels them as ambiguous, and alludes to the need of more context in order to disambiguate. One plausible cause for this could be the use of a prompt with the word "ambiguous" but, again, a lack of determinism can be noted. 

Another interesting experiment we have conducted is trying to confuse ChatGPT using erroneous pairs of lemma-definitions. Following the previous prompt model in which we input the two definition, what would happen if those are wrong in some way? A first test is using wrong definitions of the lemma that appears in the sentence, and there ChatGPT detects our trap (query \ref{ex2.3} from Appendix B.2). A second test is a sentence where wrong definitions of a lemma that does not appear in the sentence are used, and were the definitions match the lemma in the sentence. Here ChatGPT detects the trap again (query \ref{ex2.4}). Lastly, every occurrence of the word is replaced, and ChatGPT gets a bit confused, but brilliantly corrects the input again (query \ref{ex2.5}).

In conclusion, ChatGPT is remarkably good at classifying meaning of polysemous words in non-ambiguous sentences, even when the definitions are not in the context. In ambiguous sentences ChatGPT does not seem to perform as good as for homonymy, but results improve with better prompting.

\subsection{Syntactic Ambiguity}

In this section we analyze the five sentences from Appendix C.1, and from each one of them:

\begin{itemize}
    \item We ask ChatGPT about the ambiguity of the sentence with the following prompt: \textbf{Is the sentence "[sentence]" ambiguous?}
    \item In some cases, we ask ChatGPT how it interprets the sentence using the following prompt: \textbf{How do you interpret the sentence "[sentence]"?}
    \item In some cases, we ask ChatGPT about the ambiguous clause. The prompt model is: \textbf{In the sentence "[sentence]" is the clause "[clause]" ambiguous?}
\end{itemize}

As in previous steps, in first place we empirically analyse a typical sentence, and ask ChatGPT how it interprets it (query \ref{ex3.1} in Appendix C.2). Here ChatGPT does not see any ambiguity in this well-known ambiguous case. If we ask if this is an ambiguous sentence, ChatGPT does not see the ambiguity (query \ref{ex3.2}). A more precise question, as in query \ref{ex3.3}, where we ask about the meaning of the ambiguous clause, neither works. If we insist (query \ref{ex3.4}), the model even gets confused, and finds ambiguity in "saw", but not in our target, again because of an too-easily detected polysemy. In the end (query \ref{ex3.5} from Appendix C.2) we partly achieved our goal by questioning the model previous outputs. 

But, how and when does ChatGPT learn? We do the same initial question about ambiguity the same day but from another account, and ChatGPT fails again as it did with our initial query. However, if we repeat the question after a week and from a different account, then the model gets it right, as seen in query \ref{ex3.6}. It was not easy, but eventually ChatGPT recognized this typical example. It seems that the model learns within an user session, and then after some days that knowledge becomes general. 

Does ChatGPT recognize other syntactically ambiguous examples? In the sentence "I am happy you are proud, and so is Mary" the second clause can mean that "Mary" is either happy or proud. In query \ref{ex3.7} from Appendix C.2 we ask the model how it interprets this sentence, and ChatGPT again does not see any ambiguity in it. If the specifically ask if the sentence is ambiguous (query \ref{ex3.8}), ChatGPT does not see any ambiguity. If we ask about that specific clause and its ambiguity (query \ref{ex3.9}), the model neither detects ambiguity. Eventually, ChatGPT amends when we insist (query \ref{ex3.10}). A better prompt and human feedback help on this task. Also, when removing the comma before the clause (query \ref{ex3.11}), ChatGPT suddenly detects ambiguity, even though the sentence becomes syntactically wrong:

Then, what does happen with the "I watch her duck" example used by \citet{jurafsky20}? Here ChatGPT clearly detects the ambiguity (query \ref{ex3.12}), but it only sees two out of five different interpretations that \citet{jurafsky20} explain for this sentence. 

Additionally we tried some more examples. Is the sentence "They are cooking apples"\footnote{\url{https://mgmasoncreative.com/2014/08/15/syntactic-ambiguity-english-language-at-its-funniest/}} ambiguous? ChatGPT correctly identifies it in query 13. 

\begin{table*}
\centering
\begin{tabular}{cccccc}
\hline
{} & \textbf{Ambiguity type} & \textbf{True Positive} & \textbf{True Negative}& \textbf{False Positive}& \textbf{False Negative}\\
\hline
\verb|| & {Homonymy} & {3} & {2}& {7}& {1}\\
 \\\hline
 \hline
\verb|| & {Polysemy} & {0} & {10}& {0}& {2}\\
 \\\hline
  \hline
\verb|| & {Syntactic} & {2} & {0}& {0}& {3}\\
 \\\hline
  \hline
\verb|| & {Semantic} & {1} & {2}& {0}& {0}\\
 \\\hline
\end{tabular}
\begin{tabular}{cccc}
\hline
\end{tabular}
\caption{Ambiguity detection results}
\label{tab:accents}
\end{table*}

Another example is the sentence "I saw the dog with one eye"\footnote{\url{https://mgmasoncreative.com/2014/08/15/syntactic-ambiguity-english-language-at-its-funniest/}}. As previously pointed in queries \ref{ex3.8}-\ref{ex3.11} with the ambiguity between "happy" and "proud", ChatGPT encounters difficulties dealing with syntactic ambiguity (query \ref{ex3.14}). However, we can insist until we push it to learn about it (query \ref{ex3.15}).

Although it has been confusing at times, this analysis has allowed us to explore syntactic ambiguity which has been the hardest to detect with ChatGPT, but has pointed some key elements of how and when the model seems to learn, and how (better prompting and human feedback) we can interact with it in order to achieved better outputs and to make it learn.

ChatGPT has the potential to improve responses through human intervention thanks to intents. We perceive how these intents may internally modify the input sent to the model to optimise the actual prompt submitted. Thus allowing new types of interactions and responses without having to constantly modify the weights of the model.

\subsection{Semantic Ambiguity}

Although semantic ambiguity relates in literature with homonymy and polysemy, in this section we will focus on co-reference resolution. The following experiments have been done for every sentence from Appendix D.1: 

\begin{itemize}
    \item We analyse the ambiguity of the sentence with the following prompt: \textbf{Is the sentence "[sentence]" ambiguous?}
    \item We explore other specific prompts to analyse the reference of the ambiguous pronouns.
\end{itemize}

 In first place we ask the model if the sentence "My mother and my sister were sad after she shouted at her" is ambiguous. As seen in query \ref{ex4.1} from Appendix D.2,  The explanation is great and fully captures the logic behind co-reference resolution. Then, if we go further and ask "who shouted at who?", ChatGPT again detects the ambiguity of the pronouns "she" and "her" (query \ref{ex4.2}).  

Another specific case of co-reference is gender disambiguation. Here ambiguity can be useful to explore stereotypical gender bias \citep{blodgett20} in language models. In the example sentences proposed by \citet{zhao18} "The physicist hired the secretary because he was overwhelmed with work." and "The physicist hired the secretary because she was overwhelmed with work.", systems must be able to make correct linking predictions in pro-stereotypical scenarios (...) and anti-stereotypical scenarios" \citep{zhao18}. These pro-stereotypical and anti-stereotypical scenarios came from data from US Department of Labor statistics. 

We ask the model if they are ambiguous (queries \ref{ex4.3} and \ref{ex4.4}). ChatGPT correctly labels them as non-ambiguous. Then we ask about the gender of both people in each sentence, because this must be determined by the pronoun of the subordinate sentence, and not by statistical stereotypes. This means that even if traditionally a gender is more prone to a given job, in these sentence the pronoun makes this sentence non-ambiguous, so there is no need to draw upon that "statistical knowledge". Grammar should be above that. But this does not happen, and ChatGPT wrongly interprets it in query \ref{ex4.5} from Appendix D.3.

This means that even in a situation where there is no ambiguity, ChatGPT puts gender bias over grammar. This could naively mean that ChatGPT is misunderstanding grammar, but, as expected, if we change the gender of the pronouns, ChatGPT suddenly does not fail, as seen in query \ref{ex4.6}.

In conclusion, in the semantic field ChatGPT does some amazing job detecting ambiguous pronouns in co-reference resolution situations, but disappointingly fails putting gender bias over grammar in situations where this grammar makes a sentence non-ambiguous.

\subsection{Generation}

A good approach to understand ChatGPT conception of ambiguity could be asking to generate an ambiguous sentence. To achieve this, the prompt \textbf{Generate an ambiguous sentence}  (Appendix E.1) is used. After generation, the question if the model consider its own sentence ambiguous came up. Surprisingly, it does not look that way to ChatGPT now, and as seen in query \ref{ex5.2}, the sentence is labelled as not inherently ambiguous. This could be seen as the main issue with ChatGPT: its lack of consistence, that could be related with a volatile and questionable self-confidence. 

\subsection{Results}

Table 1 contains the results of the ambiguity detection queries. In our data samples ChatGPT achieves an accuracy of 0.6061 and an F1 of 0.48. 

For the case of ambiguity detection in sentences with homonyms, False Positives represents the higher amount, although this was due to the detection of polysemy in some of the homonyms, and ChatGPT did well apart from that. In the case of polysemy ChatGPT correctly detected all negative examples, but failed to detect the positive ones. The model did well when classifying definitions with or without context, and also performed great with erroneous queries.

Syntactic ambiguity appears to be harder to detect, but in the end, with the help of good prompting and reinforcement learning, ChatGPT did well at detection of disambiguation. In semantics, ChatGPT performed perfectly in the detection of ambiguities. Apart from that, it has some bright sports (co-reference resolution) and some weaknesses (puts gender bias over grammar in some non-ambiguous situations). In the generation task ChatGPT did well, but also revealed some of its worse issues: the lack of systematicity. Lastly, it should also be pointed that in most of the cases ChatGPT brilliantly alludes to lack of context as the key factor in disambiguation.

\section{Conclusions}

Transformer-based LLMs have achieved enormous advances in NLP, where disambiguation has always been one of the main problems. Models like ChatGPT continue to improve these issues, but there is still plenty of work to do. 

We have provided an introduction to linguistic ambiguity, its varieties and their relevance in NLP, and performed an extend empiric analysis of how ChatGPT treats them. This analysis revealed some of the model strengths and weaknesses. We also provide the data used in the experiments, and most of the model outputs as well. 

We consider this a promising starting point towards the use of ChatGPT and generative models in disambiguation tasks. Probably by the moment this paper is read some of the flaws of the model have improved and we have to move on to another defects.

\section{Limitations}

It should be noted that this work is an empiric analysis. We understand that the main limitation of our work is that ChatGPT is a living model that evolves in time, eager to keep on learning and able to track a conversation (not only a prompt). Therefore, some of the data used here and the results produced with them may vary in the future as the model evolves. It should also be pointed that this is not an open-source model that in some occasions suffers from overload. Furthermore, it is relevant to note that experiments have been carried out for a high-resource language like English, so the results may be different for languages with other linguistic features or fewer resources.

\section{Ethics statement}

We are aware of the possibilities that models like ChatGPT can imply for both industry and academic research in the future. Our intention here is to contribute to a better understanding and development of NLP in general and this model in particular, in order to improve its performance, but always promoting a responsible use.

\bibliography{anthology,custom}
\bibliographystyle{acl_natbib}

\appendixtitleon
\appendixtitletocon
\begin{appendices}
\section{Homonymy}
\label{ex0.2}

\subsection{SENTENCES}
\begin{enumerate}
    \item How does a bear bear with the pain?
    \item The man bought a chocolate bar at the bar. 
    \item Going to the right is the right choice.
    \item There is not choice left but going to the left.
    \item The pain in my back came back after a hard day of work. 
    \item The band singer was wearing a head band during concert.
    \item You should address the problem with your address.
    \item Rose sale rose during February. 
    \item Did you watch the watch my mom gave my for Christmas?
    \item Please, can you close the trash can?
    \item I will book the book presentation for tomorrow. 
    \item I bought a bar
    \item You look right
    \item I saw a band
\end{enumerate}

\subsection{QUERIES}

\begin{enumerate}
    \item \label{ex1.1} INPUT: \textit{How does a bear bear with the pain?}

    OUTPUT: \textit{Bears are able to bear with pain in much the same way as humans do - through a combination of natural defenses and behaviors...}

    \item \label{ex1.2} INPUT: \textit{Is the sentence "How does a bear bear with the pain?" ambiguous?}
    
    OUTPUT: \textit{Yes, the sentence "How does a bear bear with the pain?" is ambiguous because it can be interpreted in two different ways. One interpretation is asking about the physical process by which a bear copes with pain, while the other interpretation is asking about the emotional process by which a bear tolerates or endures pain. Without additional context or clarification, it is difficult to determine which interpretation is intended. It would be helpful to provide more information or to rephrase the sentence to make the intended meaning clearer."}

    \item \label{ex1.3} INPUT: \textit{what does "bear" mean in the sentence "How does a bear bear with the pain?"?}

    OUTPUT: \textit{In the sentence "How does a bear bear with the pain?", the word "bear" is being used as a verb. It means to tolerate or endure something difficult or unpleasant, such as pain. The sentence is asking how a bear is able to cope with or withstand the pain it is experiencing.}
    
    \item \label{ex1.4} INPUT: \textit{"In the sentence "How does a bear bear with the pain?", what does every word "bear" mean?"}

    OUTPUT: \textit{In the sentence "How does a bear bear with the pain?" there are two instances of the word "bear". The first instance of the word "bear" is a noun that refers to a large, carnivorous mammal that is native to the northern regions of North America, Europe, and Asia. The second instance of the word "bear" is a verb that means to tolerate or endure something, often something unpleasant or difficult. So, in this sentence, the first instance of the word "bear" refers to the animal, and the second instance of the word "bear" refers to the action of tolerating or enduring something.}

    \item \label{ex1.5} INPUT: \textit{Is the sentence "The man bought a chocolate bar at the bar" ambiguous?}

    OUTPUT: \textit{Yes, the sentence "The man bought a chocolate bar at the bar" is ambiguous because it is unclear which "bar" is being referred to. It could be interpreted to mean that the man bought a chocolate bar at a place where alcohol is served (a bar), or it could mean that the man bought a chocolate bar that is shaped like a bar. }

    \item \label{ex1.6} INPUT: \textit{Is the sentence "Going to the right is the right choice." ambiguous?}
    
    OUTPUT: \textit{Yes, the sentence "Going to the right is the right choice." is ambiguous. It is unclear whether the speaker is saying that physically going to the right direction is the correct choice, or if the speaker is saying that making the right decision (in general) is the correct choice.}

    \item INPUT: \label{ex1.7} \textit{Is the sentence "There is not choice left but going to the left." ambiguous?}
    
    OUTPUT: \textit{Yes, the sentence "There is not choice left but going to the left." is ambiguous because it is unclear if the speaker is saying that there is no other choice but to go to the left or if the speaker is saying that there is no choice at all and that the person must go to the left. The sentence could be clarified by adding more context or rephrasing it.}

    \item \label{ex1.8} INPUT: \textit{Is the sentence "You should address the problem with your address." ambiguous?}
    
    OUTPUT: \textit{Yes, the sentence "You should address the problem with your address." is ambiguous because it is not clear whether "address" refers to speaking about or addressing something, or the physical location where someone lives.}

    \item \label{ex1.9} INPUT: \textit{Is the sentence "rose sale rose during February" ambiguous?}
    
    OUTPUT: \textit{Yes, the sentence "Yes, the sentence "rose sale rose during February" is ambiguous. It is unclear whether the verb "rose" is referring to the sale of roses increasing or the sale of a specific rose product increasing.}

    \item \label{ex1.10} INPUT: \textit{Is the sentence "Did you watch the watch my mom gave my for Christmas?" ambiguous?}
    
    OUTPUT: \textit{Yes, the sentence "Did you watch the watch my mom gave my for Christmas?" is ambiguous. It is unclear if the speaker is asking if the listener watched the watch as in looking at the time or if they watched the watch as in observing it.}

    \item \label{ex1.11} INPUT: \textit{Is the sentence "I will book the book presentation for tomorrow." ambiguous?}
    
    OUTPUT: \textit{Yes, the sentence "I will book the book presentation for tomorrow." is ambiguous. It is unclear whether the speaker is referring to a presentation about a book or a presentation of a book.}

    \item \label{ex1.12} INPUT: \textit{Is the sentence "I bought a bar" ambiguous?}
    
    OUTPUT: \textit{The sentence "I bought a bar" is ambiguous and could mean a few different things. It could mean that the speaker purchased a establishment that serves alcohol, or it could mean that the speaker purchased a solid rectangular piece of metal or other material that is used for support or as a barrier, or it could mean that the speaker purchased a unit of measurement of pressure. Without more context, it is difficult to determine the intended meaning.}

    \item \label{ex1.13} INPUT: \textit{Is the sentence "You look right" ambiguous?}
    
    OUTPUT: \textit{Yes, the sentence "You look right" is ambiguous as it could have multiple interpretations depending on context. It could mean that the person looks correct or appropriate, or it could mean that the person is facing the right direction.}

    \item \label{ex1.14} INPUT: \textit{Is the sentence "I saw a band" ambiguous?}
    
    OUTPUT: \textit{No, the sentence "I saw a band" is not ambiguous. It clearly states that the speaker saw a group of musicians or performers, commonly referred to as a band.}

\end{enumerate}

\section{Polysemy}

\subsection{LEMMAS: SENTENCES}
\begin{enumerate}
    \item to serve: She has served on the committee for the last 15 years
    \item pretty: The house has four bedrooms, so it's pretty big.
    \item truck: The road was completely blocked by an overturned truck
    \item orange: a glass of orange juice
    \item to cheat: She found out that he'd been cheating on her
    \item beam: A laser beam scans the disc’s surface
    \item mouse: Once you get the feel of it, using a mouse is easy.
    \item to run: He's been running a restaurant since he left school
    \item date: He asked her out on a date.
    \item bank: By the time we reached the opposite bank, the boat was sinking fast
\end{enumerate}

\subsection{QUERIES}

\begin{enumerate}
    \item \label{ex2.1} INPUT: \textit{These are two definitions of "to cheat": 1. to behave in a dishonest way in order to get what you want 2. to have a secret sexual relationship with someone who is not your husband, wife, or usual sexual partner. Which of the definitions appear in the sentence "She found out that he'd been cheating on her"?}

    OUTPUT: \textit{The first definition, "to behave in a dishonest way in order to get what you want," applies in the sentence "She found out that he'd been cheating on her."}
        
    \item \label{ex2.2} INPUT: In the sentence "She found out that he’d been cheating on her" which is the definition of "cheating"? 
    
    OUTPUT: "In this sentence, "cheating" means that the person was being unfaithful to their partner by engaging in a romantic or sexual relationship with someone other than their partner. It typically implies a breach of trust or commitment in a romantic relationship."

    \item \label{ex2.3} INPUT: \textit{These are two definitions of "truck": 1. an organization where people and businesses can invest or borrow money, change it to foreign money, etc., or a building where these services are offered 2.sloping raised land, especially along the sides of a river .  Which of the definitions appear in the sentence "I was hit by a truck"?"}

    OUTPUT: \textit{In the sentence "I was hit by a truck," the word "truck" refers to a vehicle with four wheels and a large, rectangular body that is used for transporting goods or materials. This definition is commonly used in everyday language and refers to a type of motor vehicle. It does not refer to either of the definitions you provided, which are unrelated to vehicles."}

    \item \label{ex2.4} INPUT: \textit{These are two definitions of "truck": 1. an organization where people and businesses can invest or borrow money, change it to foreign money, etc., or a building where these services are offered 2.sloping raised land, especially along the sides of a river .  Which of the definitions appear in the sentence "I was in the bank this morning"?"} 

    OUTPUT: \textit{The first definition of "truck" (an organization where people and businesses can invest or borrow money, change it to foreign money, etc., or a building where these services are offered) appears in the sentence "I was in the bank this morning." The second definition of "truck" (sloping raised land, especially along the sides of a river) does not appear in the sentence.}
, 
    \item \label{ex2.5} INPUT: \textit{"These are two definitions of "university": 1. an organization where people and businesses can invest or borrow money, change it to foreign money, etc., or a building where these services are offered 2.sloping raised land, especially along the sides of a river .  Which of the definitions appear in the sentence "I was in the university this morning"?}

     OUTPUT: \textit{The second definition of "university" does not match the context of the sentence "I was in the university this morning." In this sentence, the word "university" is most likely being used to refer to an institution of higher education, where students can earn degrees and conduct research. The first definition you provided, "an organization where people and businesses can invest or borrow money, change it to foreign money, etc., or a building where these services are offered," does not match the context of the sentence.}

\end{enumerate}

\section{Syntactic ambiguity}

\subsection{SENTENCES}
\begin{enumerate}
    \item Mary saw John with a telescope
     \item I am happy your are proud, and so is mom
     \item I watch her duck 
     \item I saw the dog with one eye
     \item They are cooking apples
\end{enumerate}

\subsection{QUERIES}

\begin{enumerate}
    \item \label{ex3.1} INPUT: How do you interpret the sentence "Mary saw John with a telescope"? 
    
    OUTPUT: "To interpret the sentence "Mary saw John with a telescope" it means that Mary used a telescope to see John. The sentence could be rephrased as "Mary used a telescope to see John." It implies that Mary was able to see John from a distance using the telescope, possibly because he was too far away to be seen with the naked eye. The sentence does not specify the context or the reason why Mary saw John with a telescope, so it is not possible to provide further information without more context.
    
    \item \label{ex3.2} INPUT: \textit{Is the sentence "Mary saw John with a telescope" ambiguous?}

    OUTPUT: \textit{No, the sentence "Mary saw John with a telescope" is not ambiguous. The sentence means that Mary used a telescope to see John. It is not unclear or open to multiple interpretations.}

    \item \label{ex3.3} INPUT: \textit{What does "with a telescope" mean in that sentence?}

    OUTPUT: \textit{"In the sentence "Mary saw John with a telescope," the phrase "with a telescope" means that Mary used a telescope to see John. It indicates that the telescope was the means or instrument through which Mary saw John."}

    \item \label{ex3.4} INPUT: \textit{In the sentence "Mary saw John with a telescope" is "with a telescope" ambiguous?}

    OUTPUT: \textit{In the sentence "Mary saw John with a telescope," the phrase "with a telescope" is not ambiguous. It indicates that Mary saw John using a telescope, or looking through a telescope. The phrase "with a telescope" is not open to multiple interpretations, and it is clear that Mary was using a telescope to see John. If there were any ambiguity in the sentence, it might be about the meaning of "saw" – it could mean that Mary literally saw John with her own eyes, or it could mean that Mary learned about John's presence or location through some other means...}

    \item \label{ex3.5} INPUT: \textit{isn't then "with a telescope" ambiguous?}

    OUTPUT: \textit{It is possible that the phrase "with a telescope" could be ambiguous in some contexts, depending on the intended meaning of the sentence and the specific words used. For example, if the sentence was "John saw Mary with a telescope," the phrase "with a telescope" could potentially be ambiguous, because it could mean either that John saw Mary through a telescope, or that John saw Mary while holding or using a telescope. In this case, the phrase "with a telescope" could be open to multiple interpretations. However, in the original sentence "Mary saw John with a telescope," the phrase "with a telescope" is not ambiguous. It clearly indicates that Mary saw John while looking through a telescope, or while using a telescope to see John. In this context, there is no room for multiple interpretations of the phrase "with a telescope." The sentence is clear and unambiguous.}
, 
    \item \label{ex3.6} INPUT: \textit{Is the sentence "Mary saw John with a telescope" ambiguous?}
    
    OUTPUT: \textit{It is not necessarily ambiguous. The sentence "Mary saw John with a telescope" could be interpreted to mean that Mary saw John while using a telescope to observe him from a distance. It could also be interpreted to mean that Mary saw John while holding a telescope, but this would depend on the context in which the sentence is used. If there is no context or further information provided, the sentence could be ambiguous.}

    \item \label{ex3.7} INPUT: \textit{How do you interpret the sentence "I am happy you are proud, and so is Mary"?}

    OUTPUT: \textit{The sentence "I am happy you are proud, and so is Mary" can be interpreted to mean that the speaker is pleased that the person they are addressing is proud, and that Mary also shares this sentiment. The phrase "I am happy" suggests that the speaker is feeling positive or pleased about something, and the phrase "you are proud" suggests that the person they are addressing has a sense of pride or accomplishment. The phrase "so is Mary" indicates that Mary also feels the same way as the speaker. This sentence could be seen as expressing support or appreciation for the person's accomplishments or pride.} 

    \item \label{ex3.8} INPUT: \textit{Is the sentence "I am happy you are proud, and so is Mary" ambiguous?}

    OUTPUT: \textit{The sentence "I am happy you are proud, and so is Mary" is not ambiguous. It clearly conveys that the speaker is happy that the person they are speaking to is proud, and that Mary is also happy. The sentence uses the conjunction "and" to connect the two clauses, which indicates that they are closely related and that the second clause provides additional information about the first clause. There is no ambiguity or uncertainty in the meaning of the sentence.} 

    \item \label{ex3.9} INPUT: \textit{In the sentence "I am happy you are proud, and so is Mary" is the clause "and so is Mary" ambiguous?}

    OUTPUT: \textit{The phrase "and so is Mary" in the sentence "I am happy you are proud, and so is Mary" is not ambiguous. It clearly conveys that Mary is happy, just as the speaker is happy. The phrase "and so is Mary" is used to provide additional information about the speaker's happiness, and there is no uncertainty or ambiguity in its meaning. It is important to consider the context in which the sentence is being used, as this can help to clarify the meaning of the sentence and any potentially ambiguous phrases or clauses. In this case, the sentence is fairly straightforward and there is no need to rely on additional context to understand its meaning.}

    \item \label{ex3.10} INPUT: \textit{Don't you see that "and so is Mary" could mean that she is proud instead of happy?} 
    
    OUTPUT: \textit{It is possible that the phrase "and so is Mary" could be interpreted as meaning that Mary is proud, rather than happy, depending on the context in which the sentence is used. However, in the sentence "I am happy you are proud, and so is Mary", the most natural and straightforward interpretation of the phrase is that it means Mary is happy, not proud. The word "so" is often used to indicate that something is the case in the same way as something else. In this case, the speaker is happy that the person they are speaking to is proud, and the phrase "and so is Mary" indicates that Mary is also happy in the same way. This is the most logical and straightforward interpretation of the phrase, and it is unlikely that it would be interpreted as meaning that Mary is proud unless there was some other context that suggested this interpretation. It is always important to consider the context in which a sentence is used, as this can help to clarify the meaning of potentially ambiguous phrases or clauses. In this case, the context of the sentence makes it clear that the phrase "and so is Mary" means that Mary is happy, rather than proud.}

    \item \label{ex3.11} INPUT: \textit{Is the sentence "I am happy you are proud and so is Mary" ambiguous?}

    OUTPUT: \textit{Yes, the sentence "I am happy you are proud and so is Mary" is potentially ambiguous. Without further context, it is unclear whether the speaker is happy that the listener is proud, or whether the speaker is happy for some other reason and is simply stating that Mary is also happy. To make the meaning of the sentence clearer, it would be helpful to provide more context or to rephrase the sentence. For example, you could say "I am happy because you are proud" or "Mary and I are both happy that you are proud."}

    \item \label{ex3.12} INPUT: \textit{Is the sentence "I watched her duck" ambiguous?}
    
    OUTPUT: \textit{The sentence "I watch her duck" could be ambiguous because it could be interpreted in two different ways: "I watch her duck" - In this interpretation, "duck" is being used as a verb, meaning to lower one's head or body quickly to avoid something, such as a thrown object or a punch. In this case, the sentence would mean that the speaker is watching the other person duck to avoid something. "I watch her duck" - In this interpretation, "duck" is being used as a noun, meaning a waterbird with a broad blunt bill, short legs, and webbed feet. In this case, the sentence would mean that the speaker is watching the other person, who is a duck. Without further context, it is difficult to determine which interpretation is intended.} 

    \item \label{ex3.13} INPUT: \textit{Is the sentence "They are cooking apples" ambiguous?}

    OUTPUT: \textit{Yes, the sentence "They are cooking apples" is ambiguous because it is not clear what "they" refers to and it is not specified whether the apples are being used for cooking or if they are a type of apple that is good for cooking.} 

    \item \label{ex3.14} INPUT: \textit{Is the the sentence "I saw the dog with one eye" ambiguous?}

    OUTPUT: \textit{The sentence "I saw the dog with one eye" is not ambiguous. It means that you saw a dog that had only one eye. It is not unclear or uncertain what is being referred to in the sentence.}

    \item \label{ex3.15} INPUT: \textit{How do you interpret the sentence "I saw the dog with one eye"}

    OUTPUT: \textit{The sentence "I saw the dog with one eye" could potentially be ambiguous, as it could be interpreted in two different ways: The dog has only one eye: In this interpretation, the sentence means that the speaker saw a dog that has only one eye, either due to injury or birth defect. The speaker saw the dog using only one eye: In this interpretation, the sentence means that the speaker saw the dog while using only one eye, perhaps because they were covering or closing the other eye. Without additional context, it is not clear which interpretation is intended. However, if the speaker provides more information or clarifies their intended meaning, the ambiguity can be resolved.}

\end{enumerate}

\section{Semantic ambiguity}

\subsection{SENTENCES}
\begin{enumerate}
    \item My mother and my sister were sad after she shouted at her
    \item The physician hired the secretary because she was overwhelmed with clients 
    \item The physician hired the secretary because he was overwhelmed with clients 
\end{enumerate}

\subsection{QUERIES}

\begin{enumerate}
    \item \label{ex4.1} INPUT: \textit{Is the sentence "My mother and my sister were sad after she shouted at her" ambiguous?}

    OUTPUT: \textit{Yes, the sentence "My mother and my sister were sad after she shouted at her" is ambiguous. Without more context, it is unclear which person is being referred to as "she" and who was shouted at by "her." It could be interpreted in two different ways:
    My mother was sad after she (my mother) shouted at her (my sister).
    My sister was sad after she (my sister) shouted at her (my mother).
    To make the meaning clear, it would be helpful to add more context or to rephrase the sentence to specify which person is being referred to as "she" and who was shouted at by "her." For example:
    "After my mother shouted at my sister, both of them were sad."
    "After my sister shouted at my mother, both of them were sad."}
        
    \item \label{ex4.2} INPUT: \textit{In the sentence "My mother and my sister were sad after she shouted at her", who shouted at who?}

    OUTPUT: \textit{In the sentence "My mother and my sister were sad after she shouted at her," it is not clear who shouted at whom. The pronoun "she" could refer to either the speaker's mother or sister, and the pronoun "her" could refer to either the person who was shouted at or the person doing the shouting. It is also possible that the pronoun "she" refers to a third person who is not mentioned in the sentence. Without more context, it is not possible to accurately determine who shouted at whom."}

    \item \label{ex4.3} INPUT: \textit{Is the sentence ""The physicist hired the secretary because she was overwhelmed with work." ambiguous?} 

    OUTPUT: \textit{No, the sentence is not ambiguous. It is clear that the physicist hired the secretary to help with the work that the physicist was overwhelmed with.}

    \item \label{ex4.4} INPUT: \textit{Is the sentence ""The physicist hired the secretary because he was overwhelmed with work." ambiguous?} 

    OUTPUT: \textit{No, the sentence is not ambiguous. It clearly states that the physicist hired the secretary because he was overwhelmed with work.}

    \item \label{ex4.5} INPUT: \textit{In the sentence "The physicist hired the secretary because she was overwhelmed with work", who the pronoun "she" refers to?} 

    OUTPUT: \textit{The pronoun "she" refers to the secretary.}. 

    \item \label{ex4.6} INPUT: \textit{In the sentence "The physicist hired the secretary because she was overwhelmed with work.", who the pronoun "he" refers to?", who the pronoun "she" refers to?}

    OUTPUT: \textit{The pronoun "he" refers to the physicist.}
     
\end{enumerate}

\section{Generation}

\subsection{QUERIES}

\begin{enumerate}
    \item \label{ex5.1} INPUT: \textit{Generate an ambiguous sentence}

    OUTPUT: \textit{Sure! Here is an ambiguous sentence: "The chicken is ready to eat." This sentence could be interpreted in two different ways: The chicken has been cooked and is now ready to be eaten. The chicken is so hungry that it is ready to eat something else. Ambiguous sentences can be confusing because they can have multiple meanings, and it's not always clear which interpretation is intended.}

    \item \label{ex5.2} INPUT: \textit{Is the sentence "The chicken is ready to eat." ambiguous?}
    
    OUTPUT: \textit{The sentence "The chicken is ready to eat." is not inherently ambiguous. It is a simple statement that the chicken is prepared and ready to be consumed as food.
    However, the sentence could potentially be ambiguous depending on the context in which it is used. For example, if the sentence is used in a conversation about a chicken that is being cooked, it might be interpreted to mean that the chicken has finished cooking and is now ready to be eaten. However, if the sentence is used in a different context, such as a conversation about a live chicken that is being kept as a pet, it could potentially be interpreted as meaning that the chicken is being offered as food to be eaten by someone or something else. Overall, the potential for ambiguity in the sentence "The chicken is ready to eat." would depend on the context in which it is used and the knowledge and assumptions of the person or people listening or reading the sentence.}

\end{enumerate}
                                
\end{appendices}

\end{document}